\documentclass[11pt,a4paper]{article}
\usepackage[hyperref]{eacl2021}
\usepackage{times}
\usepackage{latexsym}

\usepackage{microtype}

\aclfinalcopy %
\def\aclpaperid{1247} %

\usepackage{amsfonts}
\usepackage{bm}
\usepackage{array,amsmath}
\usepackage{amssymb}
\usepackage{dsfont}
\usepackage{float}
\usepackage{graphicx}
\usepackage{algorithm}
\usepackage{algorithmicx}
\usepackage{algpseudocode}
\usepackage{pgf,tikz}
\usepackage{mathrsfs}
\usepackage{multirow}
\usepackage{booktabs}
\usetikzlibrary{arrows}
\usepackage{threeparttable}
\usepackage[inline]{enumitem}
\usepackage{blindtext}

\def\figref#1{Figure~\ref{fig:#1}}
\def\figlabel#1{\label{fig:#1}\label{p:#1}}

\def\tabref#1{Table~\ref{tab:#1}}
\def\tablabel#1{\label{tab:#1}\label{p:#1}}

\def\eqref#1{Eq.~\ref{eqn:#1}}

\def\nlang{53}

\newcounter{notecounter}
\newcommand{\enotesoff}{\long\gdef\enote##1##2{}}
\newcommand{\enoteson}{\long\gdef\enote##1##2{{
			\stepcounter{notecounter}
			{\large\bf
				\hspace{1cm}\arabic{notecounter} $<<<$ ##1: ##2
				$>>>$\hspace{1cm}}}}}
\enoteson
\enotesoff

\long\def\eat#1{}

\title{Multilingual LAMA: Investigating Knowledge in Multilingual Pretrained Language Models}

\author{Nora Kassner\thanks{\mbox{\ \ } Equal contribution - random order.} , Philipp Dufter$^{*}$, Hinrich Sch\"{u}tze\\
	Center for Information and Language Processing (CIS), LMU Munich, Germany\\
	{\tt \{kassner,philipp\}@cis.lmu.de}}

\date{}

\def\typed{TyQ}
\def\untyped{UnTyQ}

\begin{document}
\maketitle

\begin{abstract}
        Recently, it has been found that
        monolingual English language models can be used as knowledge
        bases. Instead of structural knowledge base
        queries,
masked sentences such as ``Paris
        is the capital of [MASK]'' are used as probes. We
        translate the established benchmarks TREx and
        GoogleRE into \nlang{} languages.
Working with
        mBERT,
        we investigate three questions.
        (i)
Can mBERT be used as a multilingual knowledge base?
        Most prior
        work only considers English. Extending research to
        multiple languages is important for  diversity
        and accessibility.
        (ii) Is mBERT's performance as knowledge base
        language-independent
or does it vary from language to language?
(iii) 
A multilingual model is trained on more text, e.g., mBERT is
trained on 104 Wikipedias. Can mBERT leverage this for 
better performance?
        We find that using mBERT as a
        knowledge base yields varying performance across languages and
        pooling predictions across languages improves
        performance. Conversely, mBERT
        exhibits a language bias; e.g., when queried in
        Italian, it tends to predict Italy as the
        country of origin.
\end{abstract}

\section{Introduction}

Pretrained language models (LMs)
\cite{peters-etal-2018-deep,howard-ruder-2018-universal,devlin-etal-2019-bert}
can be finetuned to a variety of natural language processing
(NLP) tasks and generally yield high
performance. Increasingly, these models and their generative
variants  are used to solve tasks by simple text generation,
without any finetuning \cite{brown2020language}.
This motivated research on how much knowledge is
contained in LMs: \citet{petroni2019language} used models pretrained
with masked language to answer
fill-in-the-blank templates such as ``Paris is the capital of [MASK].''

This research so far has been exclusively  on English.  In
this paper, we focus on using \emph{multilingual} pretrained
LMs as knowledge bases.
Working with
        mBERT,
        we investigate three questions.
        (i)
Can mBERT be used as a multilingual knowledge base?
        Most prior
        work only considers English. Extending research to
        multiple languages is important for  diversity
        and accessibility.
        (ii) Is mBERT's performance as knowledge base
        language-independent
        or does it vary from language to language?
        To answer these questions, we
        translate English
        datasets
        and analyze mBERT for \nlang{} languages.
(iii) 
A multilingual model is trained on more text, e.g., BERT's training data contains 
the English Wikipedia,  but mBERT is
trained on 104 Wikipedias. Can mBERT leverage this fact?
Indeed, we show that pooling across languages
helps performance.

\begin{table}
	\centering
	\scriptsize
	\def\sep{0.05cm}
	\begin{tabular}{@{\hspace{\sep}}l@{\hspace{\sep}}
			@{\hspace{\sep}}l@{\hspace{\sep}}l}
&		\textbf{Query} & \textbf{Two most frequent predictions} \\
		\midrule
en&		X was created in MASK. 
& [Japan (170), Italy (56), …]
\\
de&		X wurde in MASK erstellt.
& [Deutschland (217), Japan (70), …]
 \\
it &		X è stato creato in MASK.
& [Italia (167), Giappone (92), …]
\\
nl&		X is gemaakt in MASK.
& [Nederland (172), Italië (50), …]\\
\midrule
en & X has the position of MASK. & [bishop (468), God (68), ...] \\
de & X hat die Position MASK. & [WW (261), Ratsherr (108), ...]\\
it & X ha la posizione di MASK. & [pastore ( 289), papa (138), ...] \\
nl & X heeft de positie van MASK. & [burgemeester (400), bisschop (276) , ...]
	\end{tabular}
	\caption{Language bias when querying (\typed{})
          mBERT. Top: For an Italian
          cloze question,
          Italy is favored as country of origin. 
Bottom: There is no overlap between the top-ranked
predictions, demonstrating the influence of language -- even
though the facts are the same:
          the same set of
          triples is evaluated across languages.
           \tabref{pooling} shows that
          pooling predictions across languages addresses
          bias and improves performance. WW = ``Wirtschaftswissenschaftler''.}
	\tablabel{bias}
\end{table}

\eat{One overarching goal of multilingual language processing is to break language barriers: for us this includes creating a system that has access to all Wikipedias in all languages at once. }

In summary our contributions are:
\begin{enumerate*}[label={\textbf{\roman{*})}}]
	\item We automatically create a multilingual version of TREx and GoogleRE covering \nlang{} languages. 
	\item We use an alternative to
fill-in-the-blank querying --
ranking entities of the type required by the template
(e.g., cities) -- and show that it
is a better tool to investigate knowledge captured by
pretrained LMs.
\item We show that mBERT answers queries across languages with varying performance: it works reasonably for 21 and worse for 32 languages.

	\item \eat{We investigate the compatibility of
          entity representations across languages and }We
          give evidence that the query language
affects results: a query formulated in Italian is
more likely to produce Italian entities (see
\tabref{bias}).
	\item Pooling predictions across languages improves performance by large margins and even outperforms monolingual English BERT.

\end{enumerate*}
Code and data are available online (\url{https://github.com/norakassner/mlama}).

\section{Data}
\subsection{LAMA}

We follow the LAMA setup introduced by \citet{petroni2019language}. More specifically, we use data from TREx \cite{elsahar-etal-2018-rex} and GoogleRE.\footnote{\url{code.google.com/archive/p/relation-extraction-corpus/}} Both consist of triples of the form (object, relation, subject). The underlying idea of LAMA is to query knowledge from pretrained LMs using templates without any finetuning: the triple (Paris, capital-of, France) is queried with the template ``Paris is the capital of [MASK].'' In LAMA, TREx has 34,039 triples across 41 relations, GoogleRE 5528 triples and 3 relations. Templates for each relation have been manually created by \citet{petroni2019language}. We call all triples from TREx and GoogleRE together \emph{LAMA}.

LAMA has been found to contain many ``easy-to-guess''
triples; e.g., it is easy to guess that a person with
an Italian sounding name is born in Italy. \emph{LAMA-UHN} is a
subset of triples
that are hard to guess
introduced by \newcite{poerner2019bert}.

\subsection{Translation}
We translate both entities and templates. We use Google
Translate to
translate templates in the form ``[X] is the capital of
[Y]''. After translation, all templates were checked for
validity (i.e., whether they contain ``[X]'',
``[Y]'' exactly once) and corrected if necessary. In addition,
German,  Hindi and Japanese templates were checked by native
speakers to assess translation quality (see \tabref{comp_manual}). To translate the entity names, we used Wikidata and Google knowledge graphs. 

mBERT covers 104 languages. Google Translate covers 77 of
these. Wikidata and Google Knowledge Graph do not
provide entity translations for all languages and not all entities are contained in the knowledge graphs. For English we can find a total of 37,498 triples which we use from now on. On average, 34\%
of  triples could be translated (macro average over
languages). We only consider languages with a coverage above
20\%, resulting in the final number of languages we include
in our study: 53. The macro average of translated triples
in these 53 languages
is 43\%.  \figref{langs} gives statistics. We call the translated dataset \emph{mLAMA}.

\begin{figure}
	\centering
	\includegraphics[width=\linewidth]{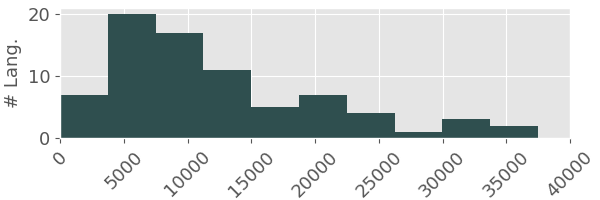}
	\caption{x-axis is the number of translated triples, y-axis the number of languages. There are 39,567 triples in the original LAMA (TREx and GoogleRE).}
	\figlabel{langs}
\end{figure}

\section{Experiments}

\subsection{Model}

We work with mBERT \cite{devlin-etal-2019-bert},\eat{\footnote{github.com/google-research/bert/blob/master/multilingual.md}} a model pretrained on the 104 largest Wikipedias. We denote mBERT queried in language x as mBERT[x]. As comparison we use the  English BERT-Base model and refer to it as BERT. In initial experiments with XLM-R \cite{conneau-etal-2020-unsupervised} we observed worse performance, similar to \newcite{jiang-etal-2020-x}. Thus, for simplicity we only report results on mBERT.

\subsection{Typed and Untyped Querying}

\citet{petroni2019language} use templates like ``Paris is the capital of [MASK]'' and give
$\arg\max_{w \in \mathcal{V}}p(w|t)$ as answer where $V$ is
the vocabulary of the LM and $p(w|t)$ is the (log-)probability
that word $w$ gets predicted in the template $t$. Thus the object of a triple must be contained in the vocabulary of the language model. This has two drawbacks: it reduces the number of triples that can be considered drastically and hinders performance comparisons across LMs with different vocabularies. We refer to this procedure as \emph{\untyped{}}.

We propose to use typed querying, \emph{\typed{}}: for each relation a candidate set $\mathcal{C}$ is created and the prediction becomes $\arg\max_{c\in \mathcal{C}}p(c|t)$. For templates like ``[X] was born in [MASK]'', we know which
entity type to expect, in this case cities. We observed
that (English-only) BERT-base predicts city names for MASK whereas mBERT predicts years for the same template. \typed{} prevents this. 

We choose as $\mathcal{C}$ the set of objects across all triples for a single relation. 
The candidate set could also be obtained from an entity typing system (e.g., \cite{yaghoobzadeh-schutze-2016-intrinsic}), but this is beyond the scope of this paper.
Variants of \typed{} have been used before \cite{Xiong2020Pretrained}.

\subsection{Singletoken vs. Multitoken Objects}

Assuming that objects are in the vocabulary
\cite{petroni2019language} is a restrictive assumption, even
more in the multilingual case as e.g., ``Hamburg'' is in the mBERT vocabulary, but 
French ``Hambourg'' is tokenized to [``Ham",
``\#\#bourg'']. We consider multitoken objects by including multiple [MASK] tokens in the templates. For both \typed{} and \untyped{} we compute the score that a multitoken object is predicted by taking the average of the log probabilities for its individual tokens. 

Given a template $t$ (e.g., ``[X] was born in [Y].'') let $t_1$ be the template with one mask token, (i.e., ``[X] was born in [MASK].'') and $t_k$ be the template with $k$ mask tokens (i.e., ``[X] was born in [MASK] [MASK] \dots [MASK].''). We denote the log probability that the token $w \in V$ is predicted at $i$th mask token as $p(m_i = w | t_k)$, where $V$ is the vocabulary of the LM.
To compute $p(e|t)$ for an entity $e$ that is tokenized into $l$ tokens $\epsilon_1, \epsilon_2, \dots, \epsilon_l$ we simply average the log probabilities across tokens:
$$p(e|t) = \frac{1}{l} \sum_{i=1}^{l} p(m_i = \epsilon_i | t_l).$$
If $k$ is the maximum number of tokens of any entity $e\in \mathcal{C}$ gets split into, we consider all templates $t_1$, \dots, $t_k$, with $C$ being the candidate set. The prediction is then the word with the highest average log probability across all templates $t_1$, \dots, $t_k$.

Note that for \untyped{} the space of possible predictions is $V\times V \times \dots \times V$ whereas for \typed{} it is the candidate set $\mathcal{C}$.

\subsection{Evaluation}
We compute precision at one  for each relation, i.e.,
$1/|T| \sum_{t \in T}\mathds{1}\{\hat{t}_{object} =
t_{object}\}$ where $T$ is the set of all triples and
$\hat{t}_{object}$ is
the object predicted by \typed{} or \untyped{}.
Note that $T$ is different for each language.
Our final measure (p1) is then
the 
precision at one averaged over relations (i.e., macro
average). Results for multiple languages are the macro average p1 across languages.

\section{Results and Discussion}

We first investigate \typed{} and \untyped{} and find that \typed{}  is better suited for investigating 
knowledge in LMs. After exploring the translation quality,
 we use \typed{} on mLAMA and
observe rather stable performance for 21 and poor
performance for 32 languages. When investigating the
languages more closely, we find that prediction results
highly depend on the language. Finally, we validate our
initial hypothesis that mBERT can leverage its
multilinguality by pooling predictions: pooling indeed
performs better.

\subsection{\untyped{} vs. \typed{}}

\eat{\begin{table*}
	\centering
	\footnotesize
	\begin{tabular}{c|c|c|c|c|c|c}
		&\multicolumn{3}{c}{ standard evaluation} &  \multicolumn{3}{c}{ candidate set evaluation} \\\hline
		& automatic & grammar & category & automatic & grammar & category \\
		\hline
	\end{tabular}
	\caption{Effect of manual template modification on standard LAMA and candidate set based evaluation. Templates are modified to correct grammatical mistakes due to the automatic translation (grammar) and are paraphrased to enable correct object category (category). We find that manual template modification has a small effect on the standard evaluation and no effect on the candidate set based evaluation. \tahttps://github.com/google-research/bert/blob/master/multilingual.mdblabel{comp_manual}}
\end{table*}
}

 \begin{figure}
 	\centering
	\includegraphics[width=\linewidth]{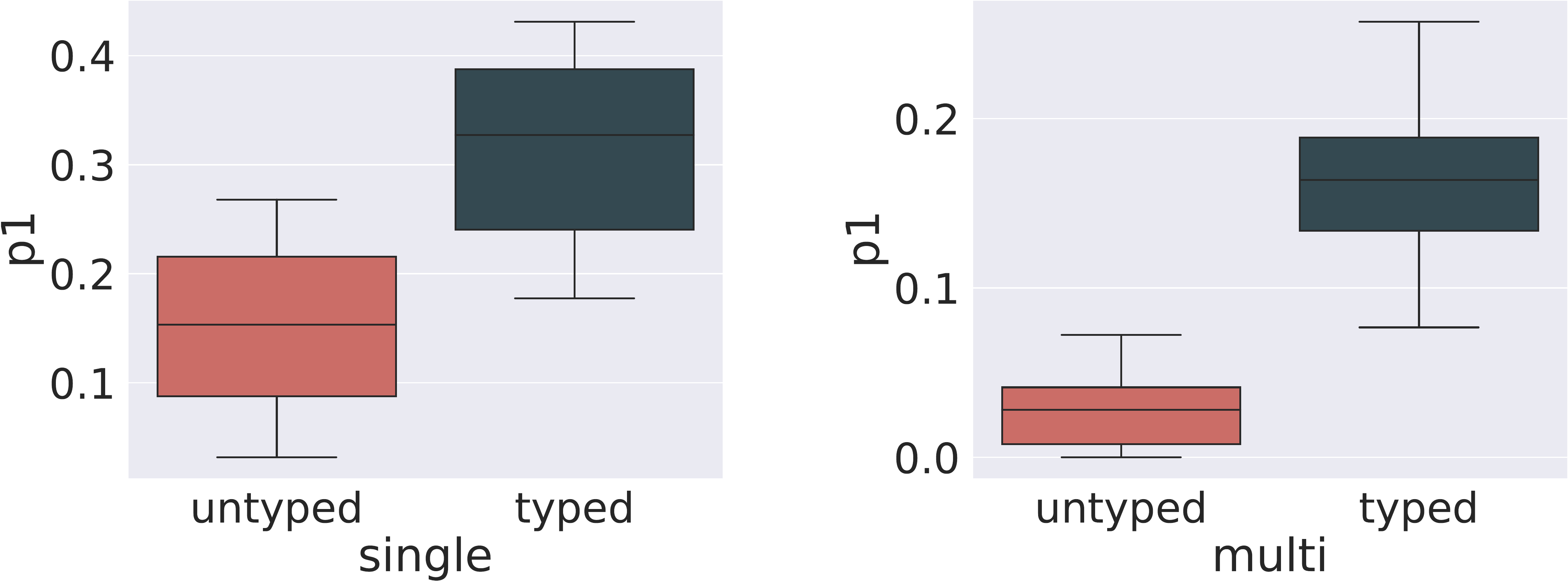}
	\caption{\figlabel{robustness}
		Distribution of p1 scores for 53 languages in \untyped{} vs.\ \typed{}.
		Left: singletoken (object $=$ 1 token). Right:  multitoken (object $>$ 1 token).}
\end{figure}

\figref{robustness} shows the distribution of p1 scores for single and multitoken objects. As expected,
\typed{} works better, both for single and multitoken objects.
\eat{Manual template tuning or automatic
	template mining is needed for good performance. 
	Template tuning for 53 languages is
	beyond the scope of this paper. Typing querying
	-allows us to isolate core knowledge performance in
	-our evaluation.}
With
\untyped{},
performance not only depends on
the model's knowledge, but on at least three extraneous factors: \textbf{(i)} Does the model understand the type constraints of
the template (e.g., in ``X is the capital of Y'', Y must be a
country)?
\textbf{(ii)} How ``fluent'' a substitution is an object under
linguistic constraints (e.g., morphology) that can be viewed as orthogonal to knowledge? Many English templates cannot be translated
into a single template in  many
languages, e.g., ``in X'' (with X a country)
has different translations in French: ``\`a Chypre'', ``au
Mexique'',
``en
Inde''. But the LAMA setup requires a single template. By
enforcing the type, we reduce the number of errors that are
due to surface fluency\eat{(e.g., what words can occur after
``en'' in French)}. \textbf{(iii)} The inadequacy of the original LAMA setup for
multitoken answers. \figref{robustness} (right) shows that the original
\untyped{} struggles with multitokens (mean p1 .03
vs.\ .17 for \typed{}).

\emph{Overall, \typed{} allows us to focus the evaluation on the core
question: what knowledge is contained in LMs? From now on, we
report numbers in the \typed{} setting.}

\begin{table}
	\centering
	\scriptsize
	\begin{tabular}{r|r|rr|rr}
		& \multicolumn{1}{c}{machine} & \multicolumn{2}{c}{manually} & \multicolumn{2}{c}{manually} \\
		& \multicolumn{1}{c}{translated} & \multicolumn{2}{c}{corrected} & \multicolumn{2}{c}{paraphrased}\\ \midrule
		de & 18.1 & 19.4 & (6)  & 20.9 & (18)\\
		hi & 5.4 & 6.2 &  (14) & 6.2 &(1) \\
		ja & 0.4 &  0.4 &  (14) &  0.7 & (5)
	\end{tabular}
	\caption{Effect of manual template modification on \untyped{}. Shown is p1, number of templates modified (in brackets). Templates are modified to correct mistakes from machine translation
		and paraphrased to achieve the correct object type. Manual template correction has a small effect on \untyped{}. \tablabel{comp_manual}}
\end{table}

Manual template tuning or automatic
template mining \cite{jiang2020can} has been investigated in the literature to approach the typing problem.
We had native speakers check templates for German, Hindi and Japanese, correct  mistakes
in the automatic translation and paraphrase the template to obtain predictions with the correct type.
\eat{Automatic templates correction was required for 6 (de), 14 (hi), 14 (ja) out of the 41 relation templates.
18  (de) and 1 (hi) and 5 (ja) paraphrased relations
improved performance. }
\tabref{comp_manual} shows that 
corrections do not yield strong improvements. We conclude that template modifications are not an effective solution for the typing problem.

 \begin{figure*}
 	\centering
  \includegraphics[width=\linewidth]{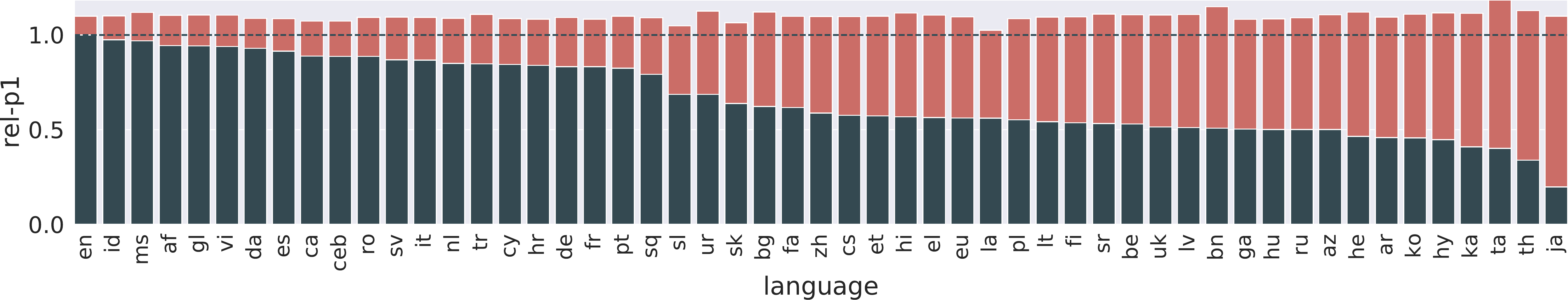}
  \caption{\figlabel{langauge_plot} p1 of
BERT (red) vs
    mBERT[x] (blue) divided by p1 of mBERT[en] on the same set of triples in each
    language x. mBERT captures less factual knowledge than
    monolingual English BERT. While performance is
    reasonable for 21 languages, it
is below 60\% for 32 languages.
Dashed line is rel-p1 of mBERT[en] (by definition equal to 1.0). Performance of BERT varies slightly as the set of triples is different for each language. Note that the Wikipedia of Cebuano (ceb) consists mostly of machine translated articles.}
\end{figure*}

\subsection{Translation Quality}

Contemporaneous work by \newcite{jiang-etal-2020-x} provides
manual translations of LAMA templates for 23 languages respecting grammatical
gender and
inflection constraints. We evaluate our machine translated
templates by comparing performance on a common subset of 14
languages using \typed{} querying on the TREx subset. Surprisingly, we
find a performance difference of 1 percentage points (0.23
vs.\ 0.24, p1 averaged over languages) in favor of the machine translated
templates.
This indicates
that the machine translated templates in combination
with \typed{} exhibit comparable performance but come with the benefit of larger language coverage (53 vs. 23 languages).

\subsection{Multilingual Performance}
In mLAMA, not all triples are available in all
languages. Thus absolute numbers are not comparable across
languages and we adopt a relative performance comparison: we
report p1 of a model-language combination divided by p1 of
mBERT's performance in English (mBERT[en]) on the exact same set of
triples and call this \emph{rel-p1}. A rel-p1 score of 0.5
for mBERT[fi] means that p1 of mBERT on Finnish is half of
mBERT[en]'s performance on the same triples. rel-p1 of English BERT is usually greater than 1 as monolingual BERT tends to outperform mBERT[en].

\figref{langauge_plot} shows that
mBERT performs reasonably well for 21 languages, but for 32
languages rel-p1 is less than 0.6 (i.e., their p1 is 60\% of
English's p1). We conclude that mBERT does not exhibit a
stable performance across languages. The variable
performance (from 20\% to almost 100\% rel-p1)
indicates that mBERT has no common representation for, say,
``Paris'' across languages, i.e., mBERT representations are language-dependent.

 \subsection{Bias} 

If mBERT captured knowledge independent of language, we should get similar answers across languages for the same relation. However, \tabref{bias}
shows that mBERT exhibits language-specific biases;
e.g., when queried in
        Italian, it tends to predict Italy as the
        country of origin. This effect occurs for several relations: \tabref{biastable} in the supplementary presents data for ten
         relations and four languages.

\subsection{Pooling}
We investigate  pooling of predictions across languages by picking the object predicted by the majority of languages. \tabref{pooling} shows  that pooled mBERT outperforms mBERT[en] by 6 percentage points on LAMA, presumably in part because
the language-specific bias is eliminated. mBERT[pooled]  even outperforms BERT by 3 percentage points on LAMA-UHN. This indicates that mBERT can leverage the fact that it is trained on 104 Wikipedias vs.\ just one and even outperforms the much stronger model BERT.

\begin{table}
	\centering
	\footnotesize
	\begin{tabular}{l |c|c}
		& LAMA & LAMA-UHN\\
		\hline
		BERT &38.5 &  29.0\\
		mBERT[en] & 35.0 & 25.7  \\
		mBERT[pooled] & \textbf{41.1} & \textbf{32.1} \\
	\end{tabular}
	\caption{p1 for BERT, mBERT queried in English, mBERT pooled on LAMA and LAMA-UHN. \tablabel{pooling}}
\end{table}

\section{Related Work}

\citet{petroni2019language} first asked the question: can
pretrained LMs function as knowledge bases? Subsequent
analyses focused on different aspects, such as negation
\cite{kassner-schutze-2020-negated}, easy to guess names
\cite{poerner2019bert}, integrating adapters
\cite{wang2020k} or finding alternatives to a
``fill-in-the-blank'' approach with single-token answers
\cite{bouraoui2019inducing,heinzerling2020language,jiang2020can}. Other work combines pretrained LM with information
retrieval
\cite{guu2020realm,lewis2020pre,Izacard2020LeveragingPR, kassner2020bert,petroni2020context}.
None of this work addresses languages other than English.

Multilingual models like
mBERT \cite{devlin-etal-2019-bert}\eat{\footnote{https://github.com/google-research/bert/blob/master/multilingual.md}}
and XLM-R \cite{conneau-etal-2020-unsupervised} 
perform well for zero-shot crosslingual transfer
\cite{hu2020xtreme}. However, we are not aware of any prior work that 
 analyzed to what degree pretrained multilingual models
can be used as knowledge bases. There are many multilingual
question answering datasets such as XQuAD
\cite{artetxe-etal-2020-cross}, TiDy \cite{clark2020tydi},
MKQA \cite{longpre2020mkqa} and MLQA \cite{lewis-etal-2020-mlqa}. Usually, multilingual models are
finetuned to solve such  tasks. Our goal is not to
improve question answering or create an alternative
multilingual question answering dataset, but instead to
investigate which knowledge is contained in
pretrained multilingual LMs without any kind of
supervised finetuning.

There is a range of alternative multilingual knowledge bases
that could be used for evaluation. Those include ConceptNet
\cite{speer2017conceptnet} or BabelNet
\cite{navigli2010babelnet}. We decided to provide a
translated versions of TREx and GoogleRE for the sake of
comparability across languages. By translating manually
created templates and entities we can ensure comparability
across languages. This is not possible for crowd-sourced
databases like ConceptNet.

In contemporaneous work,  \newcite{jiang-etal-2020-x} create
and investigate a multilingual version of LAMA. They provide
human template translations for 23 languages, propose
several methods for multitoken decoding and code-switching,
and experiment with a number of PLMs. In contrast to their work, we investigate typed querying, focus on comparabiliy and pooling across languages, and explore language biases.

\section{Conclusion}

We presented mLAMA, a dataset to investigate knowledge
in language models (LMs) in a multilingual setting covering 53
languages. While our results suggest that correct entities
can be retrieved for many languages, there is a clear
performance gap between English and, e.g., Japanese and
Thai. This suggests  that mBERT is not storing 
entity knowledge in a language-independent way. Experiments
investigating
language bias confirm this
finding. We hope that this paper and the dataset we publish
will stimulate  research on investigating knowledge in LMs
\emph{multilingually} rather than just in English.

\section*{Acknowledgements}
This work was supported
by the European Research Council (\# 740516) and the German
Federal Ministry of Education and Research (BMBF) under Grant
No. 01IS18036A. The authors of this work take full responsibility for its content. 
The second author was supported by the Bavarian research institute for digital transformation (bidt) through their fellowship program. 
We thank Yannick Couzini\'e and Karan Tiwana for correcting the Japanese and Hindi templates. 
We thank the anonymous reviewers for valuable comments.

\bibliography{anthology,eacl2021}
\bibliographystyle{acl_natbib}
\newpage

\appendix

\newpage

\begin{figure}[H]
	\centering
	\includegraphics[width=\linewidth]{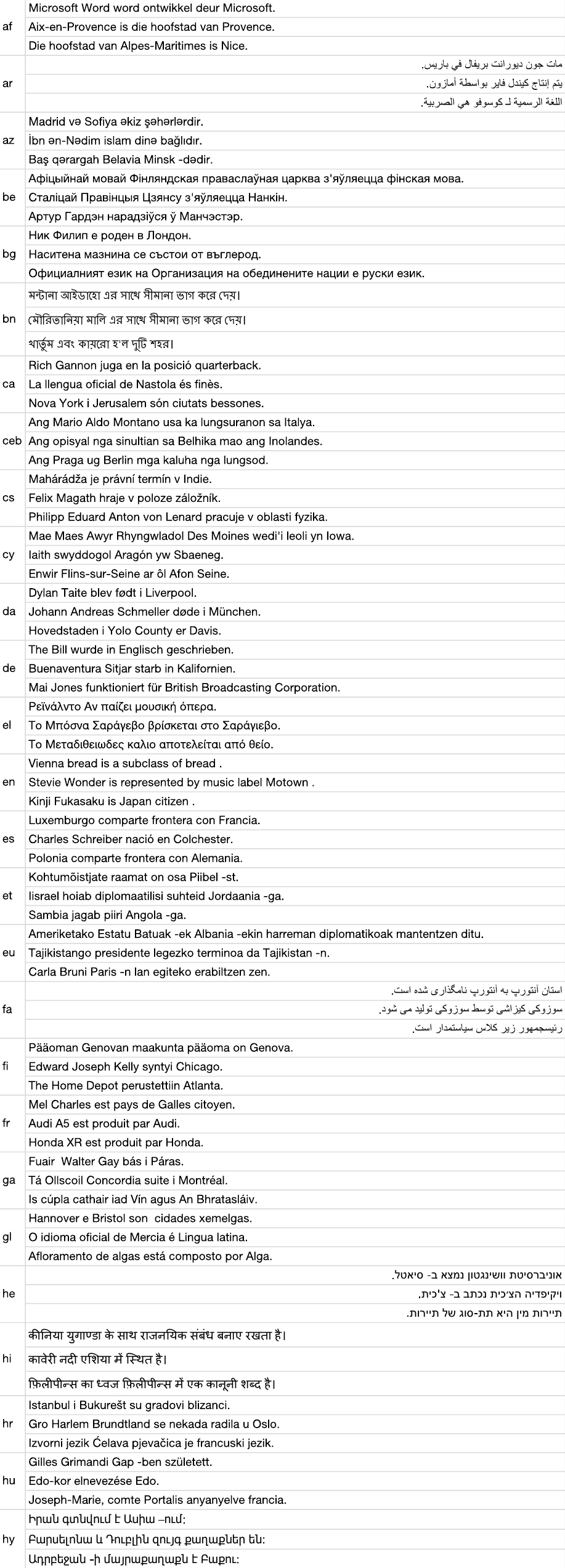}
	\caption{\tablabel{examples1} Three randomly sampled data entries from mLAMA per language. Due to the automatic generation of the dataset not all of them are fully correct.}
\end{figure}

\begin{figure}[H]
	\centering
	\includegraphics[width=\linewidth]{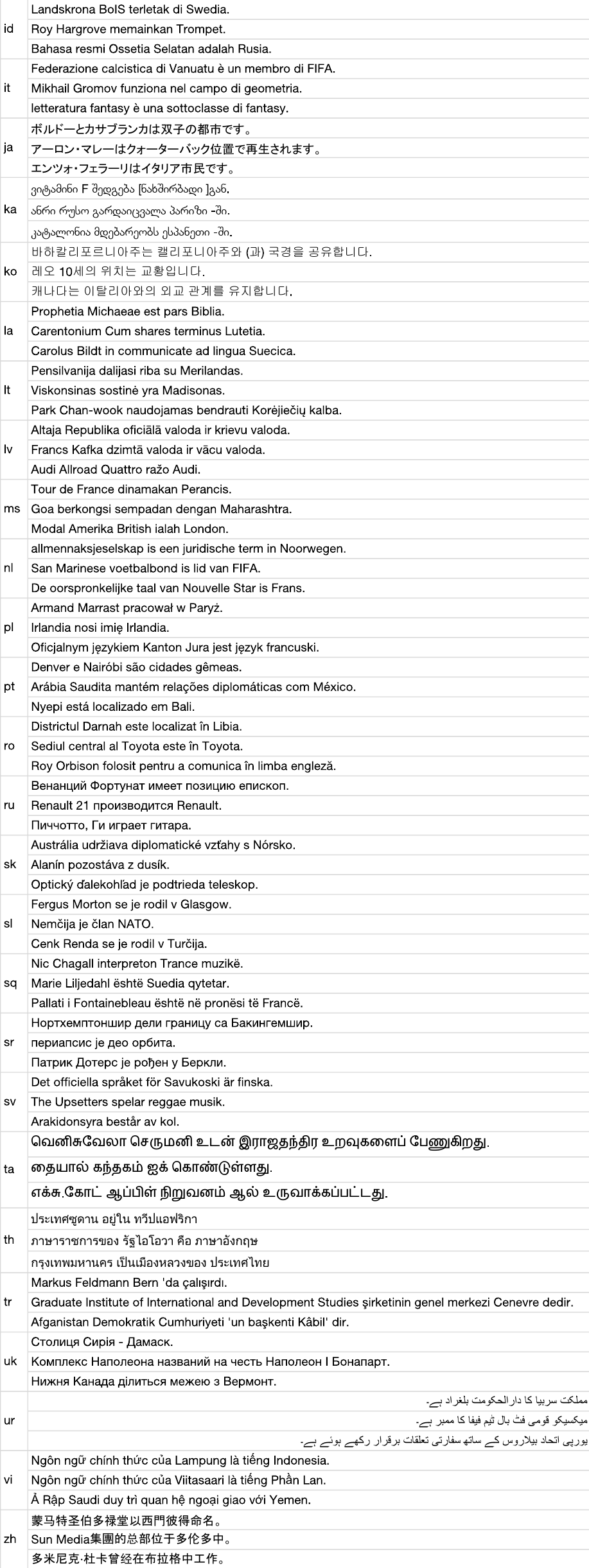}
	\caption{\tablabel{examples2} Data samples continued.}
\end{figure}

\begin{table*}[t]
	\centering
	\tiny
	\def\symmsep{0.02cm}
	\begin{tabular}{@{\hspace{\symmsep}}l@{\hspace{\symmsep}}|
			@{\hspace{\symmsep}}l@{\hspace{\symmsep}}
			@{\hspace{\symmsep}}l@{\hspace{\symmsep}}
			@{\hspace{\symmsep}}l@{\hspace{\symmsep}}
			@{\hspace{\symmsep}}l@{\hspace{\symmsep}}}
		& en & de & nl & it\\
		\hline
		P495: ``[X] was created in [Y]" & Japan (170), Italy (56)& Deutschland (217), Japan (70) & Nederland (172), Italië (50)& Italia (167), Giappone (92) \\
		P101: ``[X] works in the field of [Y]" & art (205), science (135)& Kunst (384), Film (64) & psychologie (263), kunst (120)& fisiologia (168), caccia (135) \\
		P106: ``[X] is [Y] by profession"& politician (423), composer (80)& Politiker (323), Journalist (128) & politicus (339), acteur (247)& giornalista (420), giurista (257) \\
		P1001: ``[X] is a legal term in [Y]''& India (12), Germany (11)& Deutschland (36), Russland (9) & Nederland (22), België (12)& Italia (31), Germania (16) \\
		P39: ``[X] has the position of [Y]''& bishop (468), God (68) & WW (261), Ratsherr (108)& burgemeester (400), bisschop (276) & pastore ( 289), papa (138) \\
		P527 ``[X] consists of [Y]'' & sodium (125), carbon (88)& Wasserstof (398), C (49) & vet (216), aluminium (130)& calcio (165), atomo (96) \\
		P1303 ``[X] plays [Y]''& guitar (431), piano (165)& Gitarre (312), Klavier (204) & piano (581), harp (42)& arpa (188), pianoforte (139) \\
		P178 ``[X] is developed by [Y]''& Microsoft (177), IBM (55)& Microsoft (153), Apple (99) & Microsoft (200), Nintendo (69)& Microsoft (217), Apple (49) \\
		P264 ``[X] is represented by music label [Y]''& EMI (267), Swan (32)& EMI (202), Paramount Records (59) & EMI (225), Swan (50)& EMI (217), Swan (99) \\
		P463 ``[X] is a member of [Y]''& FIFA (126), NATO (33)& FIFA (118), NATO (38) & FIFA (157), WWE (16)& FIFA (121), NATO (36)
	\end{tabular}
	\caption{Most frequent object predictions (\typed{}) in
		different languages. Some relations
		exhibit language specific biases.
		WW = ``Wirtschaftswissenschaftler''.
		\tablabel{biastable}}
\end{table*}

\newpage
\section{Language Bias}

\tabref{biastable} shows the language bias for 10 relations. For each relation we aggregated the predictions across all triples and show the most common two predicted entities together with its count (in brackets). The querying language clearly affects results. The effect is drastic for relations that ask for a country (e.g., P495 or P1001). P39 yields very different results without exhibiting a clear pattern. Other relations such as P463 or P178 are rather stable.  

\section{Data Samples}

\tabref{examples1} and \tabref{examples2} show randomly sampled entries from the data.

\section{Pretraining Data}

We investigate whether performance across languages is correlated with the amount of pretraining data for each language. To this end we investigate the number of articles per language as of January 2021\footnote{\url{https://meta.wikimedia.org/wiki/List_of_Wikipedias}} and p1 for \typed{} in \figref{sizes}. We do not have access to the original pretraining data of mBERT. Thus, the number of articles we consider in the analysis might be different to the actual data used to train mBERT. 

 \begin{figure}[t]
	\includegraphics[width=\linewidth]{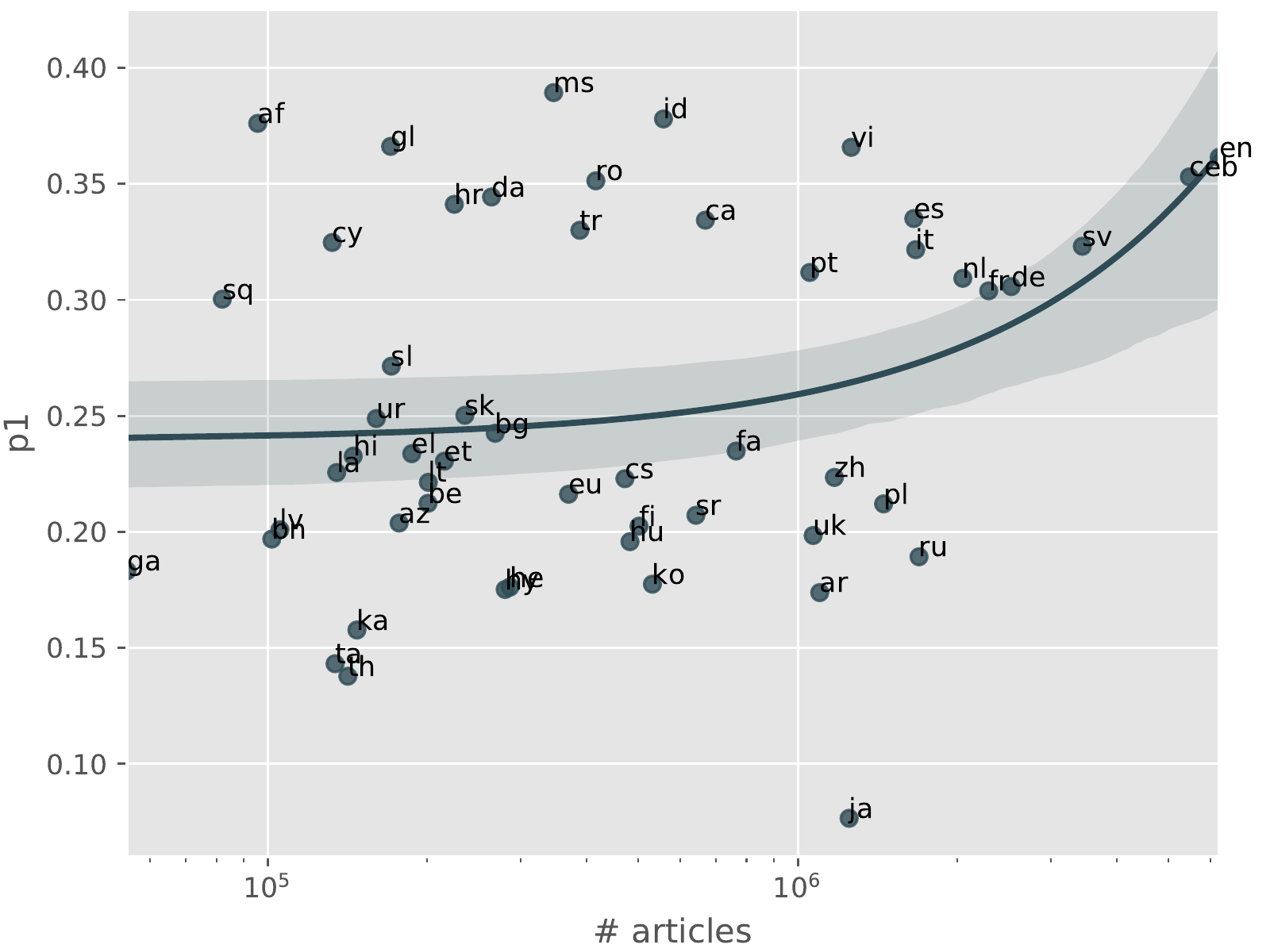}
	\caption{\figlabel{sizes}
		Scatter plot of p1 \typed{} and number of articles in the corresponding Wikipedia. There is no clear trend visible.}
\end{figure}

\eat{\section{Reproducibility Information}

Code and data used in this paper is available online (\url{https://github.com/norakassner/mlama}). We did all computations on a server with up to 40 Intel(R) Xeon(R) CPU E5-2630 v4 CPUs and 8 GeForce GTX 1080Ti GPU with 11GB memory.}

\eat{
 \begin{figure}
  \includegraphics[width=\linewidth]{images/mixing_all.png}
  \caption{\figlabel{template}
    Distribution of p1 scores for 53 languages
for translated templates (red), for translated templates
with English subjects (blue) and for English templates with
translated subjects (green).}
\end{figure}
}

\end{document}